\documentclass[runningheads]{llncs}
\usepackage{graphicx}
\usepackage{hyperref}
\hypersetup{
  colorlinks   = true, 
  urlcolor     = blue, 
  linkcolor    = blue, 
  citecolor   = blue   
}

\usepackage{color}

\usepackage{amsmath}  
\usepackage{amssymb}  

\usepackage{tabularx} 
\usepackage{booktabs} 
\usepackage{graphicx} 

\begin{document}
%
\title{Semi-Supervised Multimodal Multi-Instance Learning for Aortic Stenosis Diagnosis}
\titlerunning{Semi-supervised Multimodal MIL for Aortic Stenosis Diagnosis}
%
\author{Anonymous Submission}
\author{Zhe Huang\inst{1} \and
Xiaowei Yu\inst{2} \and
Benjamin S. Wessler \inst{3} \and
Michael C. Hughes \inst{1}}
\authorrunning{Huang et al.}
%
\institute{Tufts University, Medford, MA, USA \and
University of Texas at Arlington, Arlington, TX, USA
\and
Tufts Medical Center, Boston, MA, USA\\
}

\maketitle              
\begin{abstract}
Automated interpretation of ultrasound imaging of the heart (echocardiograms) could improve the detection and treatment of aortic stenosis (AS), a deadly heart disease. 
However, existing deep learning pipelines for assessing AS from echocardiograms have two key limitations. 
First, most methods rely on limited 2D cineloops, thereby ignoring widely available Doppler imaging that contains important complementary information about pressure gradients and blood flow abnormalities associated with AS.
Second, obtaining \emph{labeled} data is difficult. 
There are often far more unlabeled echocardiogram recordings available, but these remain underutilized by existing methods.
To overcome these limitations, we introduce Semi-supervised Multimodal Multiple-Instance Learning (SMMIL), a new deep learning framework for automatic interpretation for structural heart diseases like AS.
When deployed, SMMIL can combine information from two input modalities, spectral Dopplers and 2D cineloops, to produce a study-level AS diagnosis. During training, SMMIL can combine a smaller labeled set and an abundant unlabeled set of both modalities to improve its classifier. Experiments demonstrate that SMMIL outperforms recent alternatives at 3-level AS severity classification as well as several clinically relevant AS detection tasks. 





\end{abstract}
\section{Introduction}
\label{sec:Introduction}
Aortic stenosis (AS) is a critical degenerative heart valve condition that leads to obstructed blood flow, affecting over 12.6 million adults. With timely
diagnosis and appropriate surgical valve replacement, AS becomes a treatable condition with very low mortality~\cite{lancellotti2018outcomes}. Unfortunately, timely diagnosis of AS remains a challenge~\cite{yadgir2020global}. In current practice up to 2/3 of symptomatic AS patients may never get referred for care~\cite{tang2018contemporary,brennan2019provider}. Improved detection could help reduce the estimated 102,700 deaths caused by AS annually. 


Ultrasound (US) images of the heart, known as \emph{echocardiograms}, are considered the gold standard for AS diagnosis. Automated study-level analysis of Echocardiograms represents an opportunity for improved detection of AS. However, developing automated AS detection algorithms faces several major challenges that our present work addresses:
\begin{itemize}
\item \textbf{Mimicking expert synthesis of multiple images.} Routine US scans produce many images of the heart from diverse viewpoints. Clinicians review all available images, using only the most relevant to assess the health of the aortic valve. Developing models to emulate this intricate, multi-image review process is non-trivial as standard deep classifiers are generally designed to map a single image to a prediction. \emph{We employ a Multiple-Instance Learning framework (MIL) to interpret multiple cineloops into one comprehensive study-level prediction}, mimicking human expert diagnostic process. 


\item \textbf{Integrating across modalities.} Effective AS diagnosis in clinical settings requires the integration of information from two modalities: spectral Doppler and 2D cine series~\cite{bonow1998guidelines} (examples in Fig.~\ref{fig:workflow_diagram}). However, the fusion of these two modalities has been underexplored in existing work on AS diagnosis. \emph{We bridge this gap by designing a multimodal attention pooling mechanism that leverages the full spectrum of diagnostic information available to clinicians.}


\item \textbf{Overcoming data limitations.} The success of deep classifiers heavily relies on access to large labeled datasets, which are often scarce in medical imaging applications like our AS task. \emph{We integrate our Multimodal Multi-Instance learning framework with semi-supervised learning (SSL) to learn jointly from the labeled set and an extra unlabeled set of 5386 scans}, improving our MMIL classifier beyond what is possible with only the limited labeled set.

\end{itemize}

While prior works have made important strides toward some of these challenges, \textbf{our work tackles all three in a holistic manner}, significantly improving the state-of-the-art at detection of Aortic Stenosis from routine US scans.

\section{Related Works}
\textbf{Detecting AS with ML.} Recent years have witnessed many advances in detecting AS with machine learning. 
Holste et al.~\cite{holste2023severe} and Dai et al.~\cite{dai2023identifying} focus on detecting AS using only pre-selected PLAX cineloops from each study, while Ginsberg et al.~\cite{ginsberg2021deep} use both PLAX and PSAX. Huang et al.~\cite{huang2021new} and Wessler et al.~\cite{wessler2023automated} avoid the need of prefiltering views by 
training separate classifiers for view type and AS diagnosis, aggregating per-image predictions based on ``view relevance'' to make study-level diagnoses.
Recent works by Vaseli et al.~\cite{vaseli2023protoasnet} and Ahmadi et al.~\cite{ahmadi2023transformer} design custom neural nets tailored to 2D US, achieving state-of-the-art (SOTA). 
However, these approaches fall short by assigning diagnostic labels to each cineloop individually, rather than synthesizing one diagnosis from many inputs as a clinician does. While an overall study-level grade of AS can be produced by simply averaging instance-level predictions, not all cineloops provide proper information for diagnosis (e.g., due to image quality and content variations).
Huang et al.~\cite{huang2023detecting} tackle this issue via a MIL framework exclusive to one imaging modality (2D images), potentially missing key information about blood flow from spectral Doppler. Our present work develops a \emph{multimodal} MIL that can use both key US modalities used by clinicians.




~\\\noindent\textbf{Multiple-Instance Learning} (MIL) describes a form of weakly supervised learning where the input consists of an unordered collection of instances (a bag) along with a single label for the entire bag~\cite{dietterich1997solving}. The objective is to predict labels for unseen bags. Contemporary deep learning approaches to MIL typically follow a three-component architecture~\cite{ilse2018attention,li2021dual}: an instance representation layer that transforms each instance into a feature vector, a permutation-invariant pooling layer that aggregates these feature vectors into a bag-level representation, and an output layer that predicts the bag label based on this aggregated representation. Our proposed MMIL extends this common architecture to multimodal input.

~\\\noindent\textbf{Multimodal Fusion} is a popular strategy for modeling complex relationships between data modalities in biomedical imaging~\cite{stahlschmidt2022multimodal}. Different approaches are typically categorized into Early, Intermediate, and Late Fusion. However, the application of multimodal fusion for Echocardiograms is relatively unexplored. Our MMIL framework adopts Intermediate Fusion, effectively integrates the two key modalities in clinical practice: the 2D cineloops and the spectral Dopplers.

~\\\noindent\textbf{Semi-Supervised Learning} (SSL) addresses the scarcity of labeled data by leveraging a large pool of unlabeled data~\cite{zhu2005semi,van2020survey}. Common approaches include: \textit{Pseudo-labeling} (PL)~\cite{lee2013pseudo}, \textit{Consistency regularization}~\cite{laine2016temporal} and 
\textit{Hybrid} ~\cite{berthelot2019mixmatch,sohn2020fixmatch,huang2023fix}. Integrating SSL with MIL presents unique challenges, such as formulating appropriate unlabeled set objectives and managing the substantial GPU memory requirement.
Our approach leverages a PL-based method called curriculum labeling~\cite{cascante2021curriculum} for its simplicity and straightforward implementation.

\section{Dataset}
\label{sec:Dataset}
Our experiments use data from TMED-2~\cite{huang2021new,huang2022tmed}. It is as far as we know, the \emph{only} open-access trans-thoracic echocardiograms (TTEs) dataset with AS severity labels. TMED-2's public release contains only single-frame 2D images. We obtained permission from the dataset creators along with IRB approval (details hidden for anonymity), to access additional fully deidentified imagery from every patient-study in TMED-2, including 2D cineloops (\textbf{videos}, 20-100 per study from various anatomic views) and spectral Dopplers (\textbf{images}, 2-109 per study from various valves including the aortic valve). These data naturally form a \emph{multimodal multiple-instance learning} problem: given multiple 2D videos and Doppler images from one patient-study, predict AS severity.

For labeled data, we use TMED-2's \emph{fully-labeled set}: 599 TTE studies, with each one assigned a 3-class AS severity label indicating no AS, early AS (mild, mild-to-moderate), significant AS (moderate, severe). We repeat all experiments across the 3 distinct train/validation/test splits (n=360/119/120 patient-studies) recommended by creators~\cite{huang2022tmed}. For unlabeled data, we use TMED-2's \emph{unlabeled set}: 5486 TTE studies without any known labels.

For \textbf{external validation} of AS detection potential, we use (with creator permission and IRB approval) the temporally-external TMED-2022 set, collected after TMED-2, with 225/48/50 patient-studies each of no/early/significant AS.





\begin{figure}[!t]
\centering
\includegraphics[width=1\textwidth]{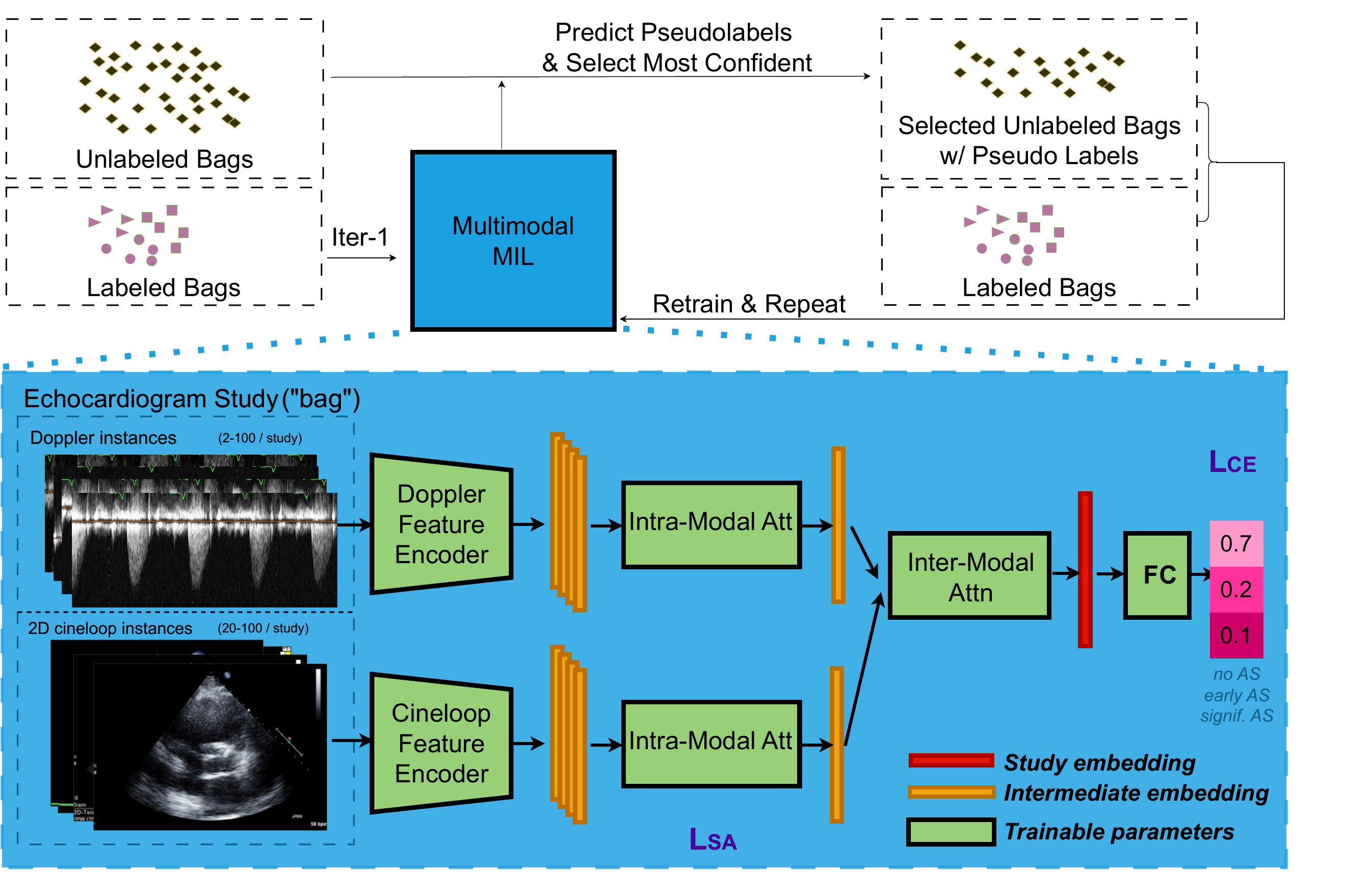}
\caption{\textbf{Overview of SMMIL}. \emph{Top:} Illustration of the SSL training workflow. In the first iteration, the model is trained on the labeled set. In all subsequent iterations, the model is trained on the union of the labeled set and the selected unlabeled subset by model from the previous iteration. This process repeats until the stopping criteria is met. \emph{Bottom:} Illustration of the Multimodal Multiple-Instance network. The network process each 2D cineloops and spectral Dopplers in a bag into feature embedding, with subsequent attention pooling operations to synthesize all information into a cohesive bag representation. Distinct feature extractors are used for 2D and spectral Dopplers branch to accommodate their unique characteristics.
}
\label{fig:workflow_diagram}
\end{figure}



\section{Methods}
\label{sec:Methods}
We now introduce our Semi-Supervised Multi-modal Multi-Instance
Learning framework (SMMIL), illustrated in
Fig~\ref{fig:workflow_diagram}.
The main components (detailed below) are the MMIL architecture and our SSL training procedure.
While our presentation is specific to AS diagnosis, our framework is general and could apply to many other multimodal multi-instance medical imaging problems.


~\\
\noindent \textbf{Problem Formulation.}
Let $D^L=\{(X_1, Y_1), \ldots, (X_N, Y_N)\}$ be a labeled training set of $N$ TTE patient-studies.
Each study, indexed by integer $i$, has a ``bag'' $X_i$ of multi-modal instances. Bag $X_i$ contains $K_i$ distinct 2D cineloops $\{x_{i1}, x_{i2}, \ldots, x_{iK_i}\}$ and $\tilde{K}_i$ distinct Dopplers $\{\tilde{x}_{i1}, \tilde{x}_{i2}, \ldots, \tilde{x}_{i\tilde{K}_i}\}$.  Each bag $X_i$ has a diagnostic label $Y_i \in \{0, 1, 2\}$ (0 = no AS, 1 =  early AS, 2 = significant AS).
Diagnosis labels for individual instances are unavailable. 
Our goal is to develop a classifier that takes as input a bag $X_i$ of multi-modal instances and accurately predicts the study-level (bag-level) label $Y_i$.


\subsection{MMIL Architecture for Multi-modal Multi-Instance Learning}

Our overall architecture builds upon previous deep attention-based MIL~\cite{ilse2018attention} and adaptations of MIL to TTE~\cite{huang2023detecting}, with key innovations in supporting \textbf{multiple data modalites} (Doppler and 2D cineloop) and \textbf{semi-supervised training.}


~\\\noindent\textbf{Instance representation layers.} 
We use distinct encoders for each modality, denoted $f$ for 2D cineloops and $\tilde{f}$ for spectral Dopplers. Both are 
row-wise feedforward neural networks that process each instance independently and identically into an $M$-dimensional vector. 
Encoding yields one vector $h_k=f(x_k)$ for each 2D cineloop $x_k$ and one vector $\tilde{h}_k = \tilde{f}(\tilde{x}_k)$ for each Doppler $\tilde{x}_k$.
Following recent work on AS detection from TTEs~\cite{ahmadi2023transformer}, we use vision transformers:
Swin Transformer-T~\cite{liu2021swin} as  $\tilde{f}$ and
Video Swin Transformer-T~\cite{liu2022video} as $f$. 

~\\\noindent\textbf{Pooling layer for Dopper Branch.} Given all per-instance vectors $\tilde{h}_k$ from all $\tilde{K}$ Dopplers in a study, we wish to map to one overall Doppler-specific representation vector $\tilde{z} \in \mathbb{R}^M$. We form $\tilde{z}$
via attention pooling inspired by ABMIL~\cite{ilse2018attention}
\begin{equation}
    \label{eq:pooling_doppler}
    \tilde{z} = \sum_{k=1}^{\tilde{K}} \tilde{a}_k \tilde{h}_k, \quad 
    \tilde{a}_k \propto \exp(\tilde{w}^\top \tanh(\tilde{U} h_k)).
\end{equation}
Here, $\tilde{w}$ and  $\tilde{U}$ are trainable parameters of suitable size, and the attention weight vector $\{ \tilde{a}_k\}_{k=1}^K$ sums to one by construction.


~\\\noindent\textbf{Pooling layer for 2D Branch.} Given all per-instance vectors $h_k$ from all 2D cineloops in a study, we use the \emph{supervised attention pooling} proposed in SAMIL~\cite{huang2023detecting} to obtain an overall 2D cineloop-specific representation $z \in \mathbb{R}^M$. 
\begin{align}
        \label{eq:pooling_2d}
        z = \sum_{k=1}^K c_k h_k, \quad c_k = \frac{a_k b_k}{\sum_{j=1}^K a_j b_j }, 
    \quad 
\begin{array}{c}
a_k \propto \exp(w_{a}^\top \tanh(U_a h_k))
\\
b_k \propto \exp(w_{b}^\top \tanh(U_b h_k))      
\end{array}
\end{align}
Here, separate attention modules produce two distinct vectors of normalized attention weights $A = \{a_1, \ldots a_K\}$ and $B = \{b_1, \ldots b_K\}$. The parameters of the first module, $U_a, w_a$, are trained in \emph{supervised} fashion as in SAMIL~\cite{huang2023detecting}, to favor PLAX and PSAX views, which are the relevant view types that clinicians use to diagnose AS. Parameters of the second module, $U_b, w_b$, are free to learn optimal attention allocation for the overall AS diagnostic task. Constructing the ultimate attention weight $c_k$ via the product $c_k \propto a_k b_k$ ensures that irrelevant views receive low weights (due to low $a_k$ values), while relevant views may receive varying attention (due to the flexibility of $b_k$).

Supervision for $A$ comes from using a pretrained view-type classifier from~\cite{huang2023detecting}, by comparing $A = \{a_1, \ldots a_K\}$ to view relevance scores $R = \{r_1, \ldots r_K\}$ obtained by renormalizing per-video predicted relevance probabilities $r(x_k)$ from the view-type classifier, indicating the likelihood of each 2D video being PLAX or PSAX. We minimize the supervised attention loss term from SAMIL~\cite{huang2023detecting}:


\begin{align}
    \label{eq:L_SA}
    \mathcal{L}_{\text{SA}}(X; w, U) = \text{KL}(R || A) = \sum_{k=1}^K r_k \log \frac{r_k}{a_k},
    \quad r_k \propto \exp(r(x_k)/\tau)
\end{align}
Here, $\tau > 0$ is a temperature hyperparameter.
Supervision for the 2D branch is possible because view-type classifiers for 2D cineloop are readily available. For spectral Doppler, off-the-shelf view classifiers or even datasets with view labels are not available. If ready in the future, our Doppler pooling branch could easily benefit from such supervision (e.g. steering toward aortic valve Dopplers).

~\\\noindent\textbf{Multimodal Fusion.} We obtain the final patient-study-level embedding $s \in \mathbb{R}^M$ via an attention-weighted averaging of 2D representation $z$ and spectral Doppler representation $\tilde{z}$:
\begin{align}
    \label{eq:pooling_multimodal}
    s = \alpha z + (1-\alpha) \tilde{z}, \quad \alpha = \frac{\eta(z)}{\eta(z) + \eta(\tilde{z})}, \qquad \eta(z) = \exp(w_{s}^\top \tanh(U_{s} z))
\end{align}
Again, learnable parameters $w_s, U_s$ allow the relative weight put on Doppler or 2D representations to change in a study-specific fashion.


~\\\noindent\textbf{Output classifier/Overall training.}
Ultimately, a linear-softmax layer maps each study-embedding $s \in \mathbb{R}^M$ to a probability vector $\rho \in \Delta^3$ indicating the chance of 3 AS severity levels (none, early, significant).
Given $N$ bag-label pairs $X_i, y_i$, we train SMMIL parameters $\theta$ (including weights for all encoders, attention pooling, and output layers) to minimize cross-entropy plus SA loss from Eq.~\eqref{eq:L_SA} with hyperparameter $\lambda > 0$:
\begin{align}
    \arg\min_{\theta} ~ \sum_{i=1}^N -\log \rho(X_i, \theta)_{Y_i} + \lambda \mathcal{L}_{\text{SA}}(X_i, \theta)
\end{align}

\subsection{Semi-supervised Training via Curriculum Labeling for MMIL}

Available labeled TTEs are not abundant (each TMED-2 training split has only 360 studies), so we pursue \emph{semi-supervised} methods to take advantage of extra unlabeled data (5400 studies).
Let $D^L$ denote the labeled set and $D^U$ denote the unlabeled set.
Fig~\ref{fig:workflow_diagram} (top) illustrates our SSL workflow, which extends a recent PL-based SSL method called curriculum-labeling (CL)~\cite{cascante2021curriculum} to our MMIL framework.
Training proceeds in several rounds. Each round trains an MMIL architecture to converge on an available set of (pseudo-)labeled data. 
In the first round, only the actual labeled set $D^L$ is used. 
In subsequent rounds, we train on the union of the labeled set $D^L$ and a \emph{subset} of the unlabeled data $D^U$ pseudo-labeled by the model from the previous round.
Pseudolabels are computed for each unlabeled bag by taking the class with maximum predicted probability; the maximum probability value itself is retained as the associated confidence. 
Selection keeps only the unlabeled bags with the highest confidence, stepping linearly to the top 20\% for round 2, the top 40\% for round 3, and so on until 100\% of unlabeled bags are selected in round 6.
To alleviate possible confirmation bias~\cite{arazo2020pseudo} commonly observed in pseudo-labeling, we train parameters $\theta$ from a random initialization at each round same as in~\cite{cascante2021curriculum}.

\section{Experiments, Results, and Analysis}
\label{sec:Experiments_and_Results}
\textbf{Implementation.} 
For each spectral Doppler, we pad the image to center around the zero-velocity line with image height spanning from -450 to 450 cm/second, then resize the image to 160 x 200.
For each 2D cineloop, we use the released 112x112 video without additional processing, taking only the first 8 frames of each cineloop as the video.
We implement our framework in PyTorch (code to be released after anonymous review). We train the model end-to-end on one 80 GB NVIDIA A100 GPU. For all experiments, we use SGD optimizer with momentum 0.9, set the batch size to 1 bag, and $\lambda$ to 10. We search several hyperparameters (learning rate in 5e-4 and 5e-5; weight decay in 1e-4 and 1e-5; temperature $\tau$ in 0.5 and 0.05) based on validation set performance.

\begin{table}[!b]
\caption{Balanced accuracy evaluation of 3-level AS severity classification on TMED-2 (left: using all available 2D instances; right: using only 2D instances with view labels).
}
\begin{tabular}{c c}
\begin{minipage}{.45\textwidth}
\resizebox{\textwidth}{!}{
    \begin{tabular}{l | rrr | r }
	    & \multicolumn{4}{c}{Bal. Acc. on Test, All 2D}\\
     Method & 1 & 2 & 3 & avg (std)\\
    \hline
    Holste et al.~\cite{holste2023severe} & 62.1 & 65.1 & 70.3 & 65.9 (3.4)
    \\
    ABMIL~\cite{ilse2018attention}                 & 58.5 & 60.4 & 61.6 & 60.2 (1.3)\\
    Set Transf.~\cite{lee2019set} & 61.0 & 62.6 & 62.6 & 62.1 (0.8)\\
    DSMIL~\cite{li2021dual}  & 60.1 & 67.6 & 73.1 & 66.9 (5.3)\\
    SAMIL~\cite{huang2023detecting}            & 72.7 & 71.6 & 73.5 & 72.6 (0.8)\\
    \hline
    SMMIL-ID & \emph{81.7} & \emph{78.8} & \emph{85.6} & \emph{82.0} (2.8)\\
    SMMIL-VD & \textbf{83.1} & \textbf{80.6} & \textbf{86.6} & \textbf{83.5} (2.5)\\
    SMMIL-V {\tiny (no doppler)} & 73.0 & 75.3 & 73.7 & 74.0 (1.0)\\
    ~MMIL-VD {\tiny (no SSL)} & 79.8 & 77.7 & 81.7 & 79.7 (1.6)\\
    \end{tabular}
}
\end{minipage}
&
\begin{minipage}{.52\textwidth}
\raisebox{1.13\height}{
\resizebox{\textwidth}{!}{
\begin{tabular}{l|rrr| r r r}
	    & \multicolumn{6}{c}{Bal. Acc. on Test, ViewLOnly 2D}\\
     Method & 1 & 2 & 3 & avg &(std)\\
    \hline
    Wessler et al.~\cite{wessler2023automated} 
    & 74.5 & 72.6 & 76.2 & 74.4 &(1.5) \\    ProtoASNet~\cite{vaseli2023protoasnet} 
    & 79.7 & --& -- & -- \\
    Ahmadi et al.~\cite{ahmadi2023transformer}
    & 83.8 & -- & -- & -- \\
    \hline
    SMMIL-ID & \emph{86.7} & \emph{82.8} & \textbf{87.7} & \emph{85.7} &(2.1)\\
    SMMIL-VD & \textbf{87.8} & \emph{82.1} & \textbf{89.6} & \textbf{86.5} &(3.2)\\
    \hline
    MedFlam~\cite{moor2023med} (zero) 
    & 50.0 & 50.0 & 50.0 & 50.0 &(0) \\
    MedFlam~\cite{moor2023med} (COT) 
    & 50.0 & 50.0 & 50.0 & 50.0 &(0) \\
    Rad~\cite{wu2023towards} (zero) 
    & 60.4 & 53.3 & 56.0 & 56.6 &(2.9) \\
    \end{tabular}
}
}
\end{minipage}
\end{tabular}
\label{tab:TMED2_BACC_Full_and_2D}
\end{table}


\textbf{Evaluation of 3-level AS Classification on TMED-2.}
We compare our SMMIL with various strong alternatives, including general MIL models and dedicated AS diagnosis models~\cite{holste2023severe} \cite{wessler2023automated} \cite{ahmadi2023transformer} \cite{vaseli2023protoasnet}. 
Comparisons to past work on TMED-2 are complicated by variations in how instances of the 2D modality were treated. First, some works use all available 2D images (\emph{All 2D}), while others examine only the 2D images/videos with associated view labels (\emph{ViewLOnly 2D}, roughly $48\%$ of all 2D). We trained and evaluated SMMIL on \emph{both} versions to ensure fair comparison. Second, past TMED-2 data releases have only provided one still frame image from each 2D instance, rather than our present focus on video. We thus report both SMMIL-ID ( \textbf{I}mage 2D instances + \textbf{D}opplers) and SMMIL-VD (\textbf{V}ideo 2D instances + \textbf{D}oppler).

Results for all methods are in Table~\ref{tab:TMED2_BACC_Full_and_2D}, where we report the balanced accuracy across the 3 splits of TMED-2.
Our SMMIL-ID pipeline achieves significant improvements (which we suggest primarily come from the Doppler modality), with video adding further modest gains.
On the harder \emph{All 2D} version, SMMIL-ID averages 82\% balanced accuracy compared to 72.6\% for the best alternative (SAMIL).
On the \emph{ViewLOnly} version, SMMIL-ID beats very recent work by \cite{ahmadi2023transformer} by 2.7 points and \cite{wessler2023automated} by over 10 points.

\textbf{Ablation.}
Two key components of SMMIL are the use of unlabeled data with SSL and the incorporation of spectral Dopplers. We assess the impact of each individual component. In Tab.~\ref{tab:TMED2_BACC_Full_and_2D} (left), we see spectral Doppler adds 9 percentage points to the average balanced accuracy, while SSL adds almost 4.





\begin{table}[!h]
\caption{AS detection tasks on Temporal Distinct Set. Showing AUROC and AUPR. 2.5th and 97.5 percentile are obtained using 5000 bootstrap samples.}
\centering
\begin{tabular}{lcccccc}
\toprule
 & \multicolumn{2}{c}{No VS Some AS} & \multicolumn{2}{c}{Early VS SigAS} & \multicolumn{2}{c}{Sig vs NoSig AS} \\
\cmidrule(lr){2-3} \cmidrule(lr){4-5} \cmidrule(lr){6-7}
Method & AUROC & AUPR & AUROC & AUPR & AUROC & AUPR \\
\midrule
W. Avg & .93(.90,.96) & .87(.80,.93) & .65(.54,.76) & .59(.46,.75) & .88(.84,.92) & .48(.35,.62) \\
ABMIL & .90(.86,.94) & .77(.68,.86) & .73(.63,.83) & .74(.60,.85) & .93(.90,.96) & .67(.53,.81) \\
DSMIL & .90(.86,.93) & .76(.66,.84) & .77(.66,.86) & .79(.67,.88) & .90(.86,.95) & .63(.48,.77) \\
SAMIL & .92(.89,.96) & .87(.81,.93) & .72(.61,.81) & .74(.61,.85) & .92(.89,.95) & .70(.56,.81) \\
\hline
SMMIL  & \textbf{.98}(.96,.99) & \textbf{.960}(.93,.98) & \textbf{.90}(.83,.96) & \textbf{.93}(.87,.97) & \textbf{.97}(.96,.99) & \textbf{.89}(.82,.95) \\
\bottomrule
\end{tabular}
\label{tab:screeningTask_FullSet}
\end{table}


\textbf{External validation.}
Tab.~\ref{tab:screeningTask_FullSet} compares our SMMIL-VD to alternatives at binary detection tasks on the external validation set. SMMIL-VD outperforms alternatives, often by a wide margin (e.g. +5 point gain on AUROC in no vs. some AS).

\textbf{Comparison to Off-the-shelf Medical Foundation Models.}
Recent efforts to build medical foundation models (MFMs)~\cite{moor2023med,wu2023towards,liu2023qilin} have shown some promise, but readiness for echocardiogram interpretation remains an open question.
We thus evaluate two SOTA MFMs, Med-Flamingo~\cite{moor2023med} and Rad-FM~\cite{wu2023towards}, on our AS diagnosis task with zero-shot, few-shot~\cite{brown2020language} and Chain-of-Thought prompting~\cite{wei2022chain}.
In Table~\ref{tab:TMED2_BACC_Full_and_2D} (right), we find that all MFMs get below 57\% balanced accuracy, barely better than random chance. This occurs despite the easier No vs Some AS binary task and substantial effort in prompt engineering. For example prompts, see Supplement Table~\ref{tab:prompts}. 
While both models do well when asked to name the body part (heart) or the imaging modality (ultrasound), our experiments on AS diagnosis suggest, unsurprisingly, that MFMs are still nascent \cite{wu2023towards} and will require further effort to succeed off-the-shelf at echo-related diagnosis.

\section{Conclusion}
We have introduced SMMIL, a deep learning solution for automated AS diagnosis using echocardiogram imaging.
We demonstrated how \emph{combining modalities} (spectral Doppler and 2D cineloop) and \emph{SSL on abundant unlabeled data} can improve AS detection. Our SMMIL can be applied to other medical tasks with multiple instances of several modalities, such as Alzheimer’s Disease detection with multi-slice MRI and clinical reports.
%
%
%
\newpage
\bibliographystyle{splncs04}
\bibliography{main.bib}

\begin{thebibliography}{10}
\providecommand{\url}[1]{\texttt{#1}}
\providecommand{\urlprefix}{URL }
\providecommand{\doi}[1]{https://doi.org/#1}

\bibitem{ahmadi2023transformer}
Ahmadi, N., Tsang, M., Gu, A., Tsang, T., Abolmaesumi, P.: Transformer-based spatio-temporal analysis for classification of aortic stenosis severity from echocardiography cine series. IEEE Transactions on Medical Imaging  (2023)

\bibitem{arazo2020pseudo}
Arazo, E., et~al.: Pseudo-labeling and confirmation bias in deep semi-supervised learning. IJCNN  (2020)

\bibitem{berthelot2019mixmatch}
Berthelot, D., Carlini, N., Goodfellow, I., Papernot, N., Oliver, A., Raffel, C.A.: Mixmatch: A holistic approach to semi-supervised learning. NeurIPS  (2019)

\bibitem{bonow1998guidelines}
Bonow, R.O., et~al.: Guidelines for the management of patients with valvular heart disease: executive summary a report of the american college of cardiology/american heart association task force on practice guidelines. Circulation  (1998)

\bibitem{brennan2019provider}
Brennan, J.M., et~al.: Provider-level variability in the treatment of patients with severe symptomatic aortic valve stenosis. Journal of the American College of Cardiology  (2019)

\bibitem{brown2020language}
Brown, T., et~al.: Language models are few-shot learners. NeurIPS  (2020)

\bibitem{cascante2021curriculum}
Cascante-Bonilla, P., Tan, F., Qi, Y., Ordonez, V.: Curriculum labeling: Revisiting pseudo-labeling for semi-supervised learning. AAAI  (2021)

\bibitem{dai2023identifying}
Dai, W., Nazzari, H., Namasivayam, M., Hung, J., Stultz, C.M.: Identifying aortic stenosis with a single parasternal long-axis video using deep learning. Journal of the American Society of Echocardiography  (2023)

\bibitem{dietterich1997solving}
Dietterich, T.G., Lathrop, R.H., Lozano-P{\'e}rez, T.: Solving the multiple instance problem with axis-parallel rectangles. Artificial intelligence  (1997)

\bibitem{ginsberg2021deep}
Ginsberg, T., Tal, R.e., Tsang, M., Macdonald, C., Dezaki, F.T., van~der Kuur, J., Luong, C., Abolmaesumi, P., Tsang, T.: Deep video networks for automatic assessment of aortic stenosis in echocardiography. Simplifying Medical Ultrasound: Second International Workshop  (2021)

\bibitem{holste2023severe}
Holste, G., Oikonomou, E.K., Mortazavi, B.J., Coppi, A., Faridi, K.F., Miller, E.J., Forrest, J.K., McNamara, R.L., Ohno-Machado, L., Yuan, N., et~al.: Severe aortic stenosis detection by deep learning applied to echocardiography. European Heart Journal  \textbf{44}(43),  4592--4604 (2023)

\bibitem{huang2021new}
Huang, Z., Long, G., Wessler, B., Hughes, M.C.: A new semi-supervised learning benchmark for classifying view and diagnosing aortic stenosis from echocardiograms. Machine Learning for Healthcare Conference  (2021)

\bibitem{huang2022tmed}
Huang, Z., Long, G., Wessler, B., Hughes, M.C.: Tmed 2: a dataset for semi-supervised classification of echocardiograms. In DataPerf: Benchmarking Data for Data-Centric AI Workshop, ICML  (2022)

\bibitem{huang2023fix}
Huang, Z., Sidhom, M.J., Wessler, B., Hughes, M.C.: Fix-a-step: Semi-supervised learning from uncurated unlabeled data. AISTATS  (2023)

\bibitem{huang2023detecting}
Huang, Z., Wessler, B.S., Hughes, M.C.: Detecting heart disease from multi-view ultrasound images via supervised attention multiple instance learning. Machine Learning for Healthcare Conference  (2023)

\bibitem{ilse2018attention}
Ilse, M., Tomczak, J., Welling, M.: Attention-based deep multiple instance learning. ICML  (2018)

\bibitem{laine2016temporal}
Laine, S., Aila, T.: Temporal ensembling for semi-supervised learning. ICLR  (2017)

\bibitem{lancellotti2018outcomes}
Lancellotti, et~al.: Outcomes of patients with asymptomatic aortic stenosis followed up in heart valve clinics. JAMA cardiology  (2018)

\bibitem{lee2013pseudo}
Lee, D.H.: Pseudo-label: The simple and efficient semi-supervised learning method for deep neural networks. Workshop on challenges in representation learning  (2013)

\bibitem{lee2019set}
Lee, J., Lee, Y., Kim, J., Kosiorek, A., Choi, S., Teh, Y.W.: Set transformer: A framework for attention-based permutation-invariant neural networks. ICML  (2019)

\bibitem{li2021dual}
Li, B., Li, Y., Eliceiri, K.W.: Dual-stream multiple instance learning network for whole slide image classification with self-supervised contrastive learning. CVPR  (2021)

\bibitem{liu2023qilin}
Liu, J., Wang, Z., Ye, Q., Chong, D., Zhou, P., Hua, Y.: Qilin-med-vl: Towards chinese large vision-language model for general healthcare. arXiv  (2023)

\bibitem{liu2021swin}
Liu, Z., Lin, Y., Cao, Y., Hu, H., Wei, Y., Zhang, Z., Lin, S., Guo, B.: Swin transformer: Hierarchical vision transformer using shifted windows. CVPR  (2021)

\bibitem{liu2022video}
Liu, Z., Ning, J., Cao, Y., Wei, Y., Zhang, Z., Lin, S., Hu, H.: Video swin transformer. CVPR  (2022)

\bibitem{moor2023med}
Moor, M., et~al.: Med-flamingo: a multimodal medical few-shot learner. Machine Learning for Health (ML4H)  (2023)

\bibitem{sohn2020fixmatch}
Sohn, K., Berthelot, D., Carlini, N., Zhang, Z., Zhang, H., Raffel, C.A., Cubuk, E.D., Kurakin, A., Li, C.L.: Fixmatch: Simplifying semi-supervised learning with consistency and confidence. NeurIPS  (2020)

\bibitem{stahlschmidt2022multimodal}
Stahlschmidt, S.R., Ulfenborg, B., Synnergren, J.: Multimodal deep learning for biomedical data fusion: a review. Briefings in Bioinformatics  (2022)

\bibitem{tang2018contemporary}
Tang, L., et~al.: Contemporary reasons and clinical outcomes for patients with severe, symptomatic aortic stenosis not undergoing aortic valve replacement. Circulation: Cardiovascular Interventions  (2018)

\bibitem{van2020survey}
Van~Engelen, J.E., Hoos, H.H.: A survey on semi-supervised learning. Machine learning  (2020)

\bibitem{vaseli2023protoasnet}
Vaseli, H., Gu, A.N., Ahmadi~Amiri, S.N., Tsang, M.Y., Fung, A., Kondori, N., Saadat, A., Abolmaesumi, P., Tsang, T.S.: Protoasnet: Dynamic prototypes for inherently interpretable and uncertainty-aware aortic stenosis classification in echocardiography. MICCAI  (2023)

\bibitem{wei2022chain}
Wei, J., Wang, X., Schuurmans, D., Bosma, M., Xia, F., Chi, E., Le, Q.V., Zhou, D., et~al.: Chain-of-thought prompting elicits reasoning in large language models. NeurIPS  (2022)

\bibitem{wessler2023automated}
Wessler, B.S., et~al.: Automated detection of aortic stenosis using machine learning. Journal of the American Society of Echocardiography  (2023)

\bibitem{wu2023towards}
Wu, C., Zhang, X., Zhang, Y., Wang, Y., Xie, W.: Towards generalist foundation model for radiology. arXiv  (2023)

\bibitem{yadgir2020global}
Yadgir, S., et~al.: Global, regional, and national burden of calcific aortic valve and degenerative mitral valve diseases, 1990--2017. Circulation  (2020)

\bibitem{zhu2005semi}
Zhu, X.J.: Semi-supervised learning literature survey. University of Wisconsin-Madison Department of Computer Sciences  (2005)

\end{thebibliography}
%





\newpage
\appendix
\begin{center}
\Large Supplementary Material
\end{center}

\newcommand{\BW}{0.32}
\setlength{\tabcolsep}{0.05cm}
\begin{figure}[h]
\begin{tabular}{r c c c }
    & Split 1 & Split 2 & Split 3
    \\
    {\rotatebox{90}{~~~~~~~~W. Avg~\cite{wessler2023automated} }}
    & 
    \includegraphics[width=\BW\textwidth]{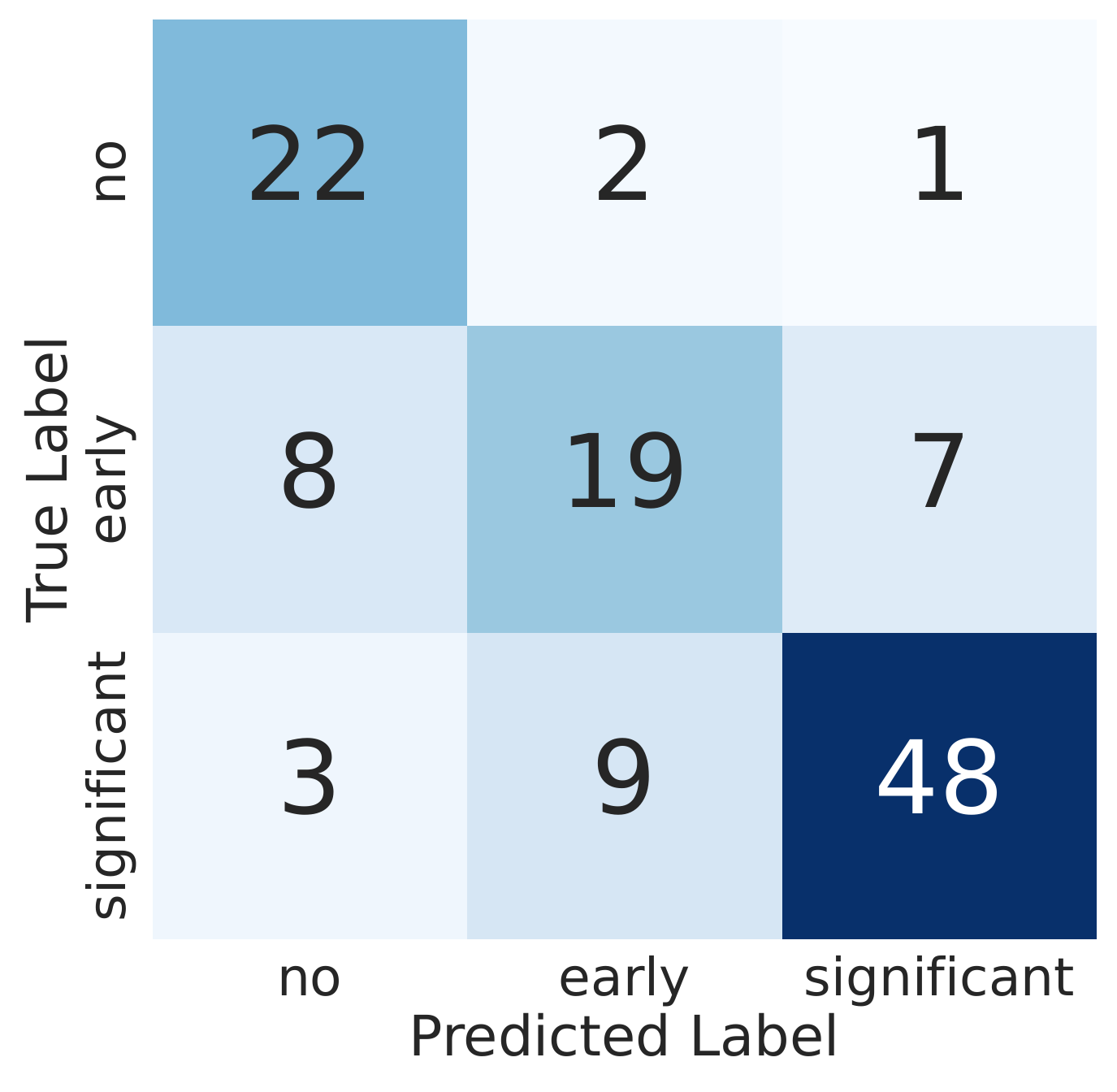}
    &
    \includegraphics[width=\BW\textwidth]{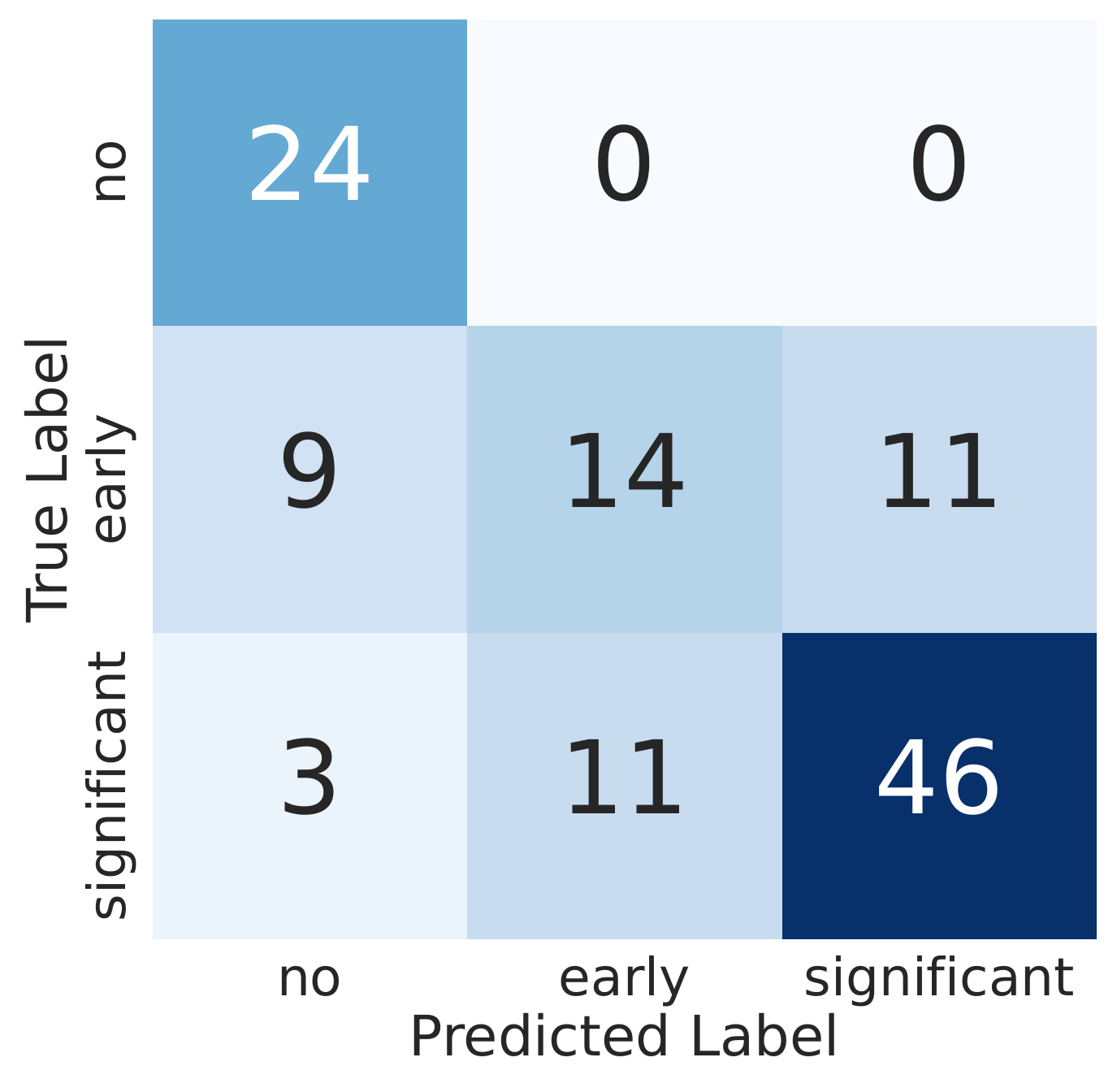}
    &
    \includegraphics[width=\BW\textwidth]{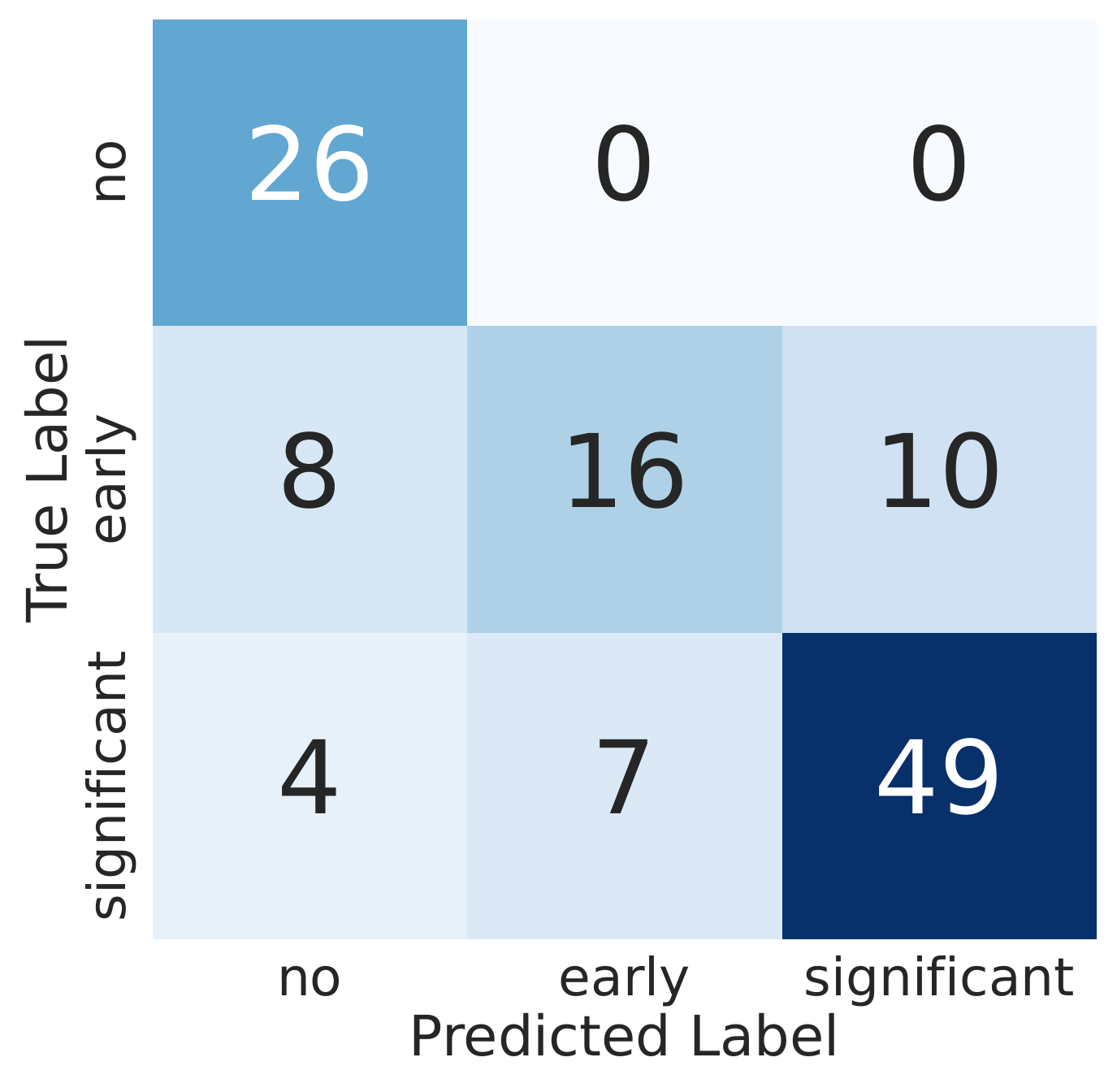}
    \\
    {\rotatebox{90}{~~~~~~~~ SMMIL (ours)}}
    & 
    \includegraphics[width=\BW\textwidth]{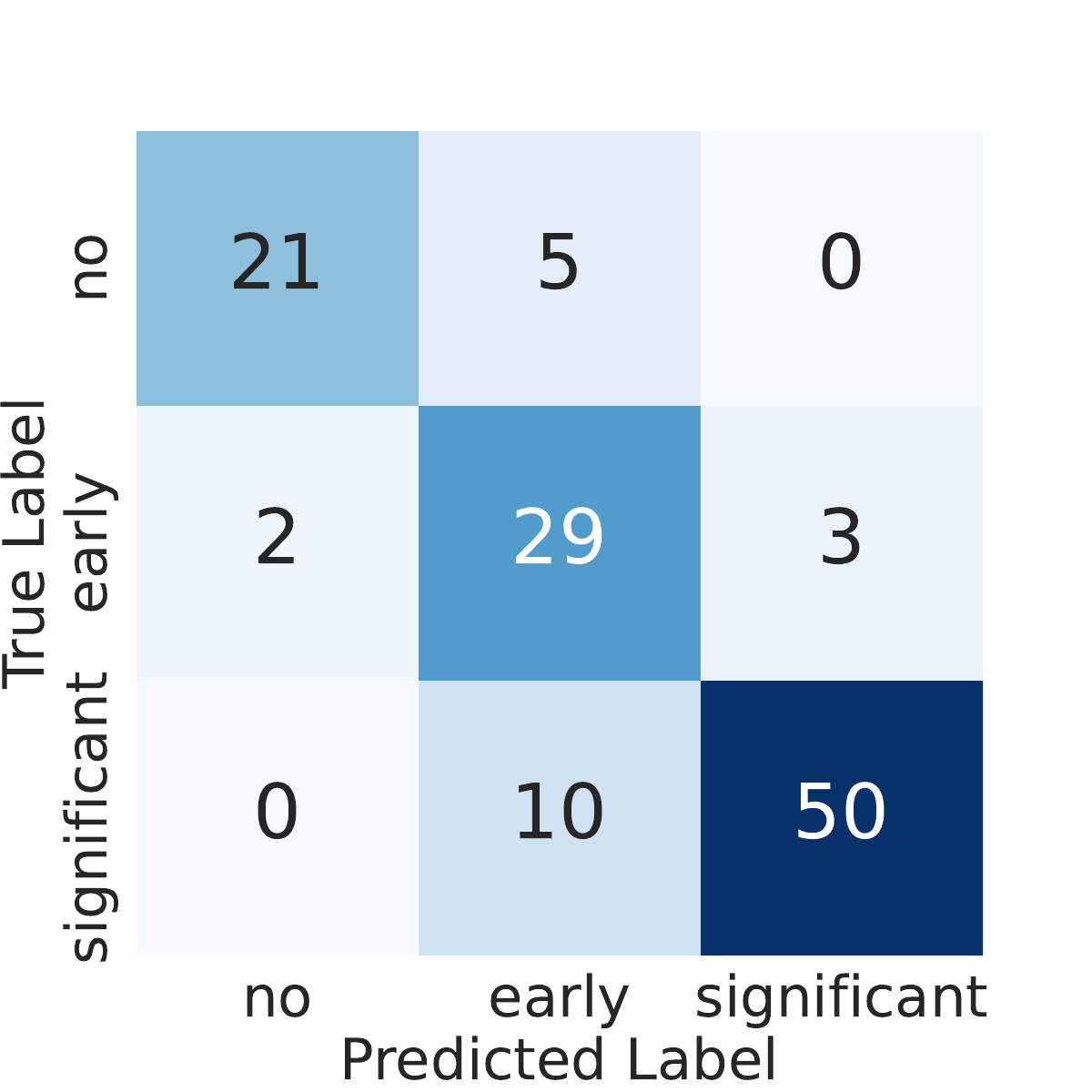}
    &
    \includegraphics[width=\BW\textwidth]{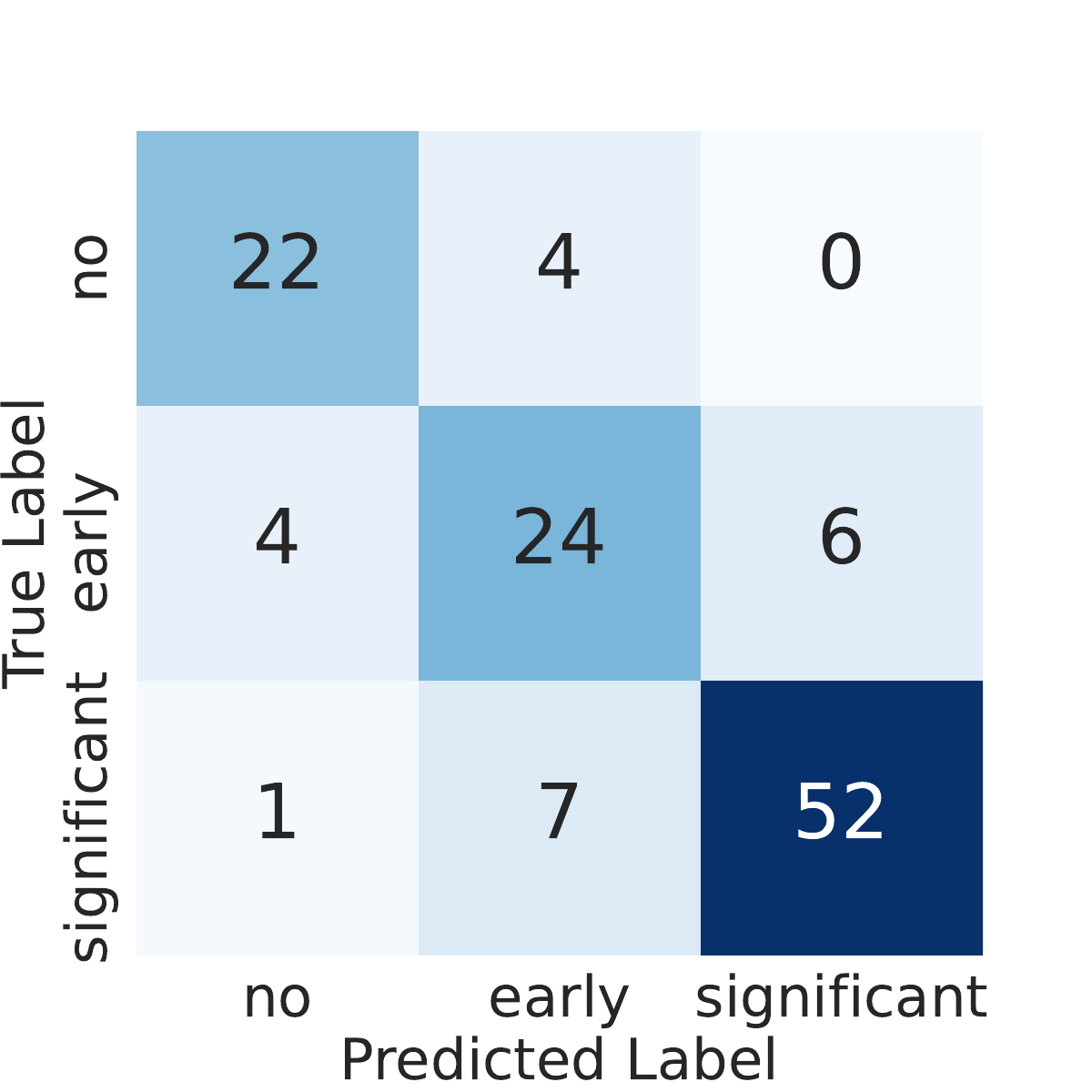}
    &
    \includegraphics[width=\BW\textwidth]{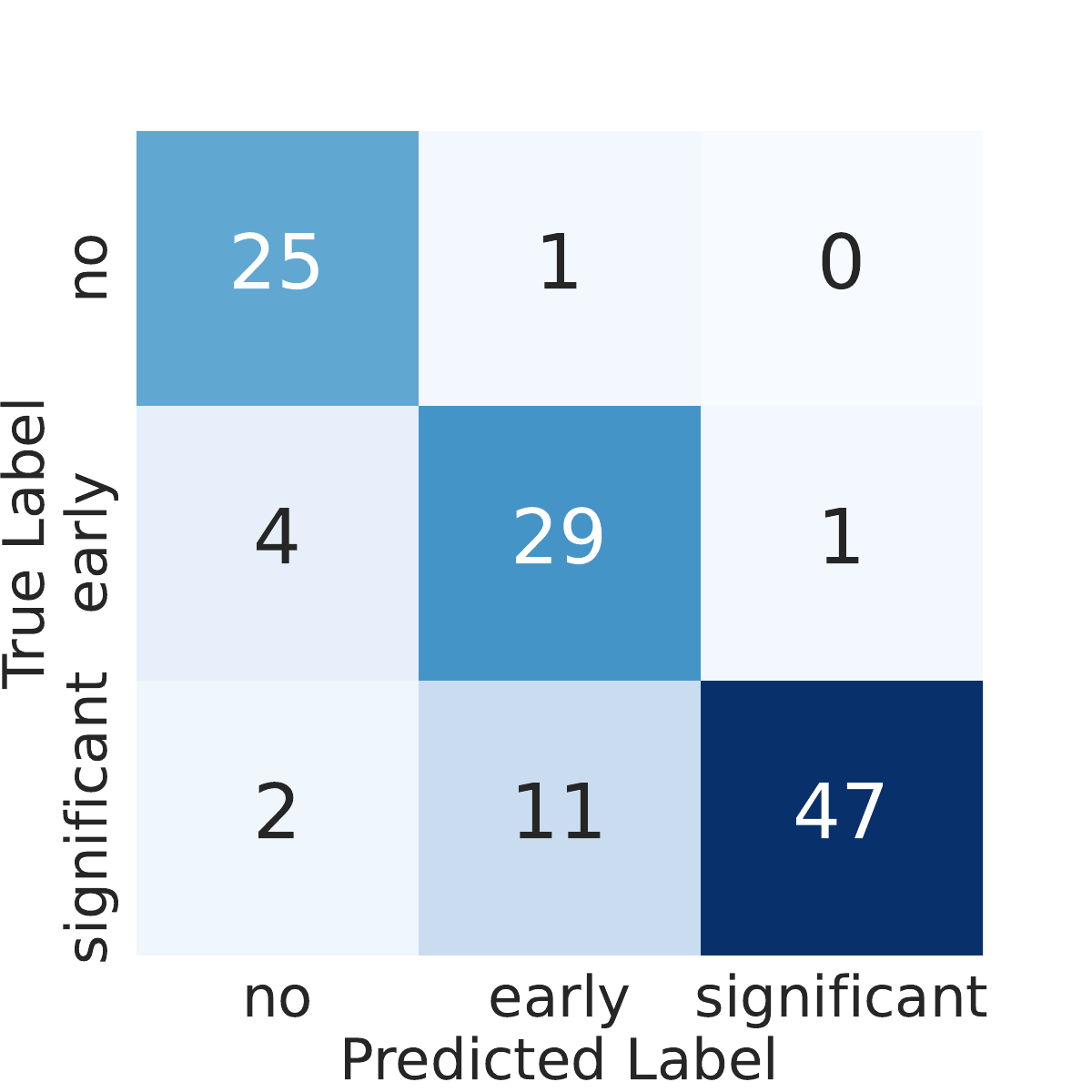}
    \end{tabular}
    \caption{Confusion Matrices for AS severity classification task across three predefined train/test splits of TMED-2.
     }
    \label{fig:confusion_matrix}
\end{figure}

\begin{figure}[h]
\begin{tabular}{c c c }
    \includegraphics[width=0.3\textwidth]{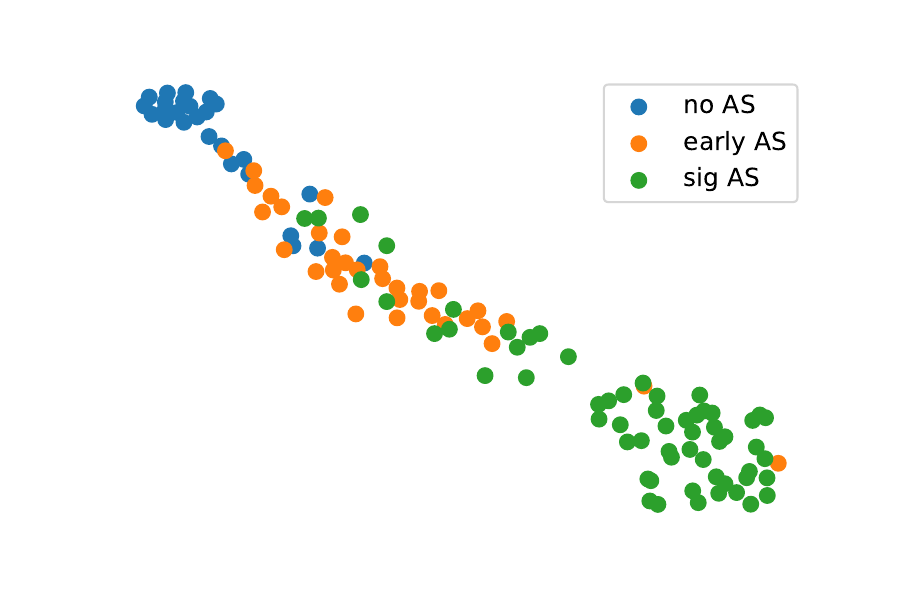}
    &
    \includegraphics[width=0.3\textwidth]{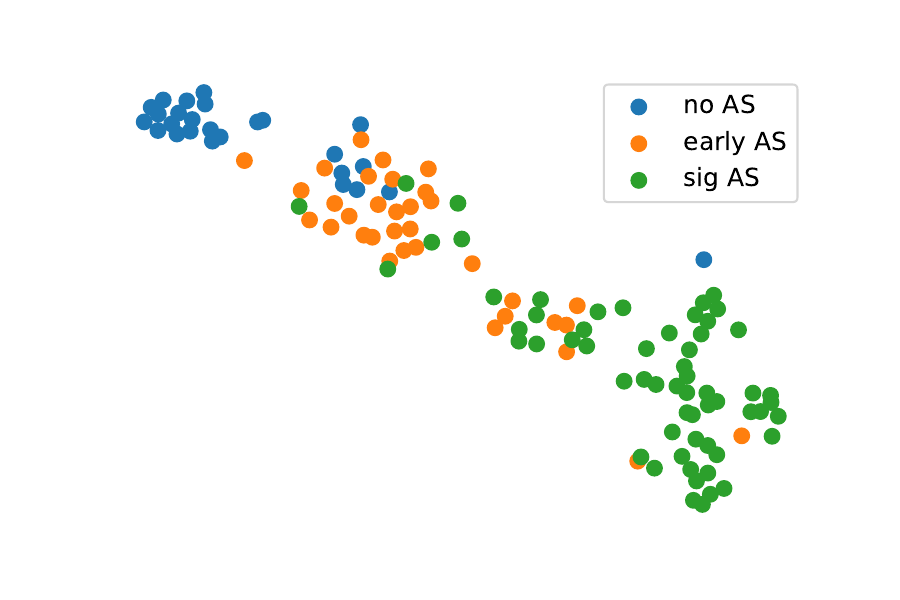}
    &
    \includegraphics[width=0.3\textwidth]{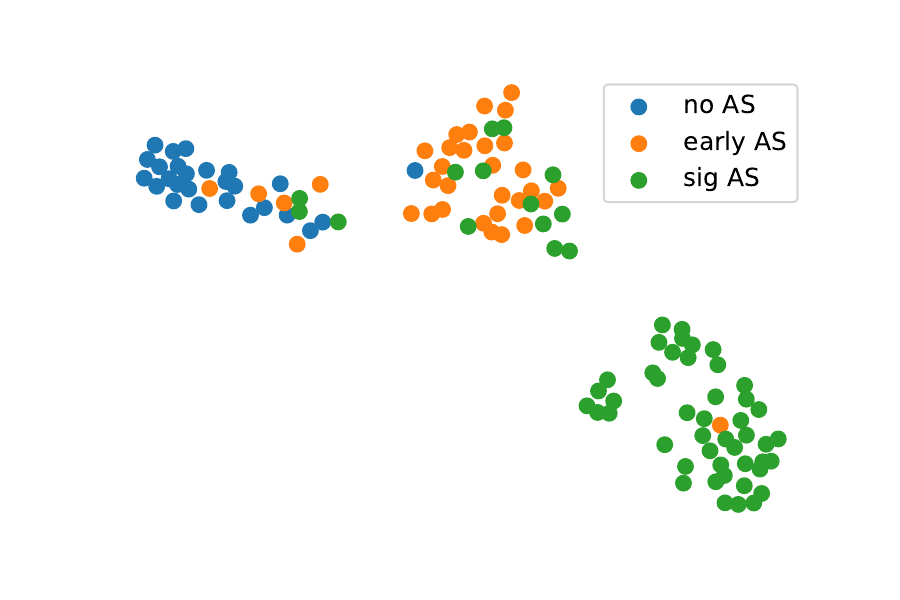}
    \end{tabular}	
    \caption{
    t-SNE visualization of TTE study representation on 3 test splits of TMED-2. The patient representations form noticeable clusters for different AS severity levels.
     }
     \label{fig:tsne_SMMIL_video_TMED2_testset}
\end{figure}
\begin{table}[ht]
\caption{Comparison of AS detection tasks performance on TMED-2. Showing average and standard deviation of the 3 test splits. Our method significantly outperform all compared baselines.}
\centering
\begin{tabular}{lcccccc}
\toprule
 & \multicolumn{2}{c}{No VS Some AS} & \multicolumn{2}{c}{Early VS SigAS} & \multicolumn{2}{c}{Sig vs NoSig AS} \\
\cmidrule(lr){2-3} \cmidrule(lr){4-5} \cmidrule(lr){6-7}
Method & AUROC & AUPR & AUROC & AUPR & AUROC & AUPR \\
\midrule
\cite{wessler2023automated} & 0.957(.018) & 0.988(.006) & 0.747(.034) & 0.785(.023) & 0.870(.015) & 0.853(.030) \\
ABMIL & 0.869(.035) & 0.958(.013) & 0.742(.048) & 0.850(.026) & 0.810(.013) & 0.823(.020) \\
DSMIL & 0.907(.034) & 0.972(.011) & 0.798(.015) & 0.892(.002) & 0.862(.007) & 0.874(.008) \\
SAMIL & 0.940(.020) & 0.982(.008) & 0.854(.026) & 0.922(.017) & 0.901(.018) & 0.910(.019) \\
\hline
SMMIL  & \textbf{0.976}(.012) & \textbf{0.993}(.004) & \textbf{0.904}(.007) & \textbf{0.943}(.008) & \textbf{0.942}(.009) & \textbf{0.930}(.010) \\
\bottomrule
\end{tabular}
\label{tab:TMED2_ScreeningTask_FullSet}
\end{table}


\begin{table}[ht]
\caption{Example of some prompts used to prompt the medical foundational models. The question string we used in COT example is ``Does this image show signs of Aortic Stenosis? Answer Choices: (a) YES (b) NO (c) Irrelevant Image''. We tried more variants than shown in this Table but wasn't able to obtain better than chance performance on Aortic Stenosis Classification task. }
\centering
\begin{tabular}{|m{0.1\linewidth}|m{0.9\linewidth}|}
\hline
zero-shot prompt& ``You are a helpful medical assistant. Please answer the question about the given image. \textless{}image\textgreater{}Question: the query question. Answer:'' \\
\hline
Default few-shot prompt& ``You are a helpful medical assistant. You are being provided with images, a question about the image and an answer. Follow the examples and answer the last question. \textless image\textgreater Question: What is/are the structure near/in the middle of the brain? Answer: pons.\textless endofchunk\textgreater\textless image\textgreater Question: Is there evidence of a right apical pneumothorax on this chest x-ray? Answer: yes.\textless endofchunk\textgreater\textless image\textgreater Question: Is/Are there air in the patient's peritoneal cavity? Answer: no.\textless endofchunk\textgreater\textless image\textgreater Question: Does the heart appear enlarged? Answer: yes.\textless endofchunk\textgreater\textless image\textgreater Question: What side are the infarcts located? Answer: bilateral.\textless endofchunk\textgreater\textless image\textgreater Question: Which image modality is this? Answer: mr flair.\textless endofchunk\textgreater\textless image\textgreater Question: Does this image show signs of Aortic Stenosis? Answer:'' \\
\hline
COT prompt & ``You are a helpful medical assistant in Echardiology, you know only Parasternal Long Axis (PLAX) view and Parasternal Short Axis (PSAX) view of the heart are relevant to diagnosing Aortic Stenosis. You are being provided with images, a question about the image, and an answer. Follow the examples and answer the last question. \textless{}image\textgreater{}Question: \{question\_str\} Answer: This is a PLAX view of the heart. This PLAX image shows signs of some Aortic Stenosis; the answer is (a) YES\textless{}image\textgreater{}Question: \{question\_str\} Answer: This is a PLAX view of the heart. However, this PLAX image does not show signs of Aortic Stenosis; the answer is (b) NO\textless{}image\textgreater{}Question: \{question\_str\} Answer: This is a PSAX view of the heart. This PSAX image shows signs of some Aortic Stenosis; the answer is (a) YES\textless{}image\textgreater{}Question: \{question\_str\} Answer: This is a PSAX view of the heart. However, this PSAX image does not show signs of Aortic Stenosis; the answer is (b) NO\textless{}image\textgreater{}Question: \{question\_str\} Answer: This is not a PLAX or PSAX view of the heart; the answer is (C) Irrelevant Image.\textless{}image\textgreater{}Question: \{question\_str\} Answer:''\\ 
\hline
\end{tabular}
\label{tab:prompts}
\end{table}

\end{document}